\documentclass[runningheads]{llncs}
\usepackage[T1]{fontenc}
\usepackage{graphicx}
\graphicspath{ {./} }

\usepackage{hyperref}
\usepackage{amsfonts}
\usepackage{amsmath}
\usepackage{siunitx}
\usepackage{adjustbox}
\usepackage{caption}
\usepackage{subcaption}
\usepackage{url}
\usepackage[flushleft]{threeparttable}

\usepackage{color}

\begin{document}
\title{SIFT-VTON: Geometric Correspondence Supervision on Cross-Attention \\ for Virtual Try-On}
\titlerunning{Geometric Correspondence Supervision for VTON}
%
\author{Kosuke Takemoto \and
Takafumi Koshinaka}
\authorrunning{K Takemoto and T Koshinaka.}
%
\institute{The Graduate School of Data Science, Yokohama City University, Kanagawa, Japan \\
\email{y245613e@yokohama-cu.ac.jp}}
\maketitle              
\renewcommand{\thefootnote}{}
\footnotetext{K. Takemoto: This author is now with ZOZO Research.}
\renewcommand{\thefootnote}{\arabic{footnote}}
\begin{abstract}
Diffusion-based virtual try-on methods achieve photorealistic synthesis
through cross-attention mechanisms that transfer garment features to target body regions.
However, these approaches rely on implicit learning of spatial correspondences,
struggling to preserve fine details such as text and illustrations.
We propose a novel approach, which we call SIFT-VTON,
that utilizes SIFT keypoint matching to provide explicit geometric guidance
for diffusion-based virtual try-on.
Our method applies domain-specific filtering to SIFT keypoint matches between garment and person images,
then converts these correspondences into spatial probability distributions
that supervise cross-attention layers during training.
This explicit supervision guides the model to learn precise spatial alignment,
concentrating attention on geometrically consistent garment regions.
Experiments on the VITON-HD dataset demonstrate significant improvements on unpaired metrics
while maintaining competitive paired reconstruction metrics.
Qualitative comparisons show superior preservation of text clarity and pattern alignment.
Attention visualizations confirm that our method produces sharply focused attention
on relevant garment details.
This work demonstrates that classical geometric correspondence methods
can effectively enhance modern diffusion models for conditional synthesis tasks.
The source code will be available at \url{https://github.com/takesukeDS/SIFT-VTON}.

\end{abstract}
\section{Introduction}
Online fashion retail has grown rapidly,
yet customers face challenges in assessing garment fit and appearance without physical try-on.
Image-based virtual try-on (VTON or VITON) aims to address this by synthesizing realistic images of people wearing desired garments,
enabling customers to visualize products on themselves before purchase.
The advent of diffusion models has significantly advanced this task,
enabling photorealistic synthesis through cross-attention mechanisms that transfer garment features to target body regions.
Despite these advances, achieving precise spatial alignment between garment patterns and body regions remains challenging,
particularly for complex textures, logos, and illustrations that require accurate geometric correspondence.

Recent methods employ diffusion models conditioned on target garment images through cross-attention mechanisms~\cite{tryondiffusion,anydoor,stableviton,ita-mdt,anydressing},
sometimes called implicit warping, producing realistic VTON images.
Despite their superiority to traditional GAN-based methods~\cite{gpvton,hrviton} in terms of image quality,
these approaches often struggle to preserve fine-grained garment features, such as logos and illustrations,
leading to noticeable artifacts in the synthesized images.

We argue that the cross-attention layers,
which determine how garment features are spatially transferred during image generation,
can benefit from explicit correspondence supervision.
Estimated warping flows, which indicate which garment pixels appear at each body location,
can be used at inference to enhance garment fidelity~\cite{ladi-vton,hyb-viton,spm-diff}.
However, these warping flows are often inaccurate,
particularly outside torso regions~\cite{hyb-viton}.
Since VTON training typically uses paired data of garments and people wearing them,
we can instead leverage accurate geometric correspondences during training.

Classical feature matching methods like Scale Invariant Feature Transform (SIFT)~\cite{sift}
have proven effective at establishing reliable geometric correspondences
across images with varying scales and orientations.
By leveraging these correspondences as supervision signals,
we can guide the attention mechanisms to learn more accurate spatial relationships
while maintaining the generative quality of diffusion models.

In this work, we propose SIFT-VTON,
a novel training approach that supervises cross-attention layers
using correspondences derived from SIFT keypoint matching.
Our method constructs spatial probability distributions from filtered SIFT matches
and uses cross-entropy loss to encourage attention weights to focus on geometrically consistent garment regions.
This explicit supervision helps the model learn precise spatial correspondences during training,
leading to improved garment spatial accuracy at inference time.

The main contributions of this paper are:
\begin{itemize}
\item A domain-specific filtering procedure for SIFT correspondences tailored to virtual try-on constraints,
ensuring reliable matches for attention supervision.
\item A cross-attention supervision method that converts geometric correspondences into probability distributions
for training diffusion models.
\item Comprehensive evaluation showing improved generation quality with better preservation
of fine-grained details such as logos and illustrations
compared to existing diffusion-based virtual try-on methods.
\end{itemize}

\section{Related Works}
Typical image-based VTON methods use a person image
and a garment image as inputs to generate an output image of the person wearing the garment.
Historically, image-based VTON is tackled with explicit warping,
where systems first explicitly transform the garment image to match the person’s pose using techniques such as
Thin-Plate Spline (TPS) or flow estimation~\cite{clothflow,hrviton,gpvton,kgi}.
The warped garment is then fed into a generative model, typically Generative Adversarial Networks (GANs),
to produce the final output image.

However, natural transformations from in-shop garments to target poses are difficult to realize with explicit warping.
Typical artifacts have been known to occur,
such as squeezing or stretching near the garment boundaries and
lack of natural wrinkles that occur when wearing clothes~\cite{hrviton,sdviton,hyb-viton}.
In some cases, accurate warping is even impossible
due to occlusions by body parts and/or limited texture information from a single garment image.

To address these limitations,
recent methods employ diffusion models with cross-attention mechanisms for implicit garment transformation~\cite{tryondiffusion,stableviton,ladi-vton,anydoor,ita-mdt,anydressing}.
Rather than explicitly warping the garment, these methods condition the denoising process on garment features through cross-attention layers,
enabling more natural synthesis of VTON images.
Garment encoders, either pre-trained models such as CLIP~\cite{clip,ladi-vton} and DINOv2~\cite{dinov2,anydoor} or jointly trained networks~\cite{tryondiffusion,stableviton},
are introduced to extract spatially informative features from the in-shop garment images.
These features are then utilized in the cross-attention layers of denoising U-Net.

Despite their advances, these methods still struggle with precise spatial alignment of garment details,
such as text, logos, and illustrations,
that require accurate geometric correspondence between garment and body regions.
To address this limitation, several methods~\cite{ladi-vton,hyb-viton,spm-diff} incorporate estimated explicit flows alongside diffusion models.
The estimated flows enable direct copying of garment regions, helping preserve fine details.
However, these flows from explicit warping methods are often inaccurate, and their errors can propagate to the final outputs.

StableVITON~\cite{stableviton} indicated that
accurate correspondences between garment features and the points where attention is focused are crucial for high-quality VTON.
This concept is related to works in text-to-image generation such as
Directed Diffusion~\cite{directed-diff} and Layout Guidance~\cite{layout-guidance},
which manipulate cross-attention maps to control specific object placement.
StableVITON proposes a loss function that regularizes cross-attention behavior.
The method computes the attention center as the weighted average of spatial locations,
where attention weights serve as the weighting coefficients,
treating this center as a point estimate of the focused region.
Their Attention Total Variation (ATV) loss encourages this center to move smoothly as query positions change.
Restricting abrupt shifts in attention centers helps maintain spatial coherence.
The loss, however, does not explicitly encourage attention to corresponding garment regions,
nor does it enforce concentration of attention weights around the computed center.

Unlike StableVITON's smooth regularization approach,
our method provides explicit supervision using geometric correspondences from SIFT keypoint matching.
This enables the model to learn precise spatial alignment while concentrating attention on geometrically consistent regions.

\section{Proposed Method}
\label{sec:method}
\begin{figure}[t]
    \centering
    \includegraphics[width=0.98\linewidth]{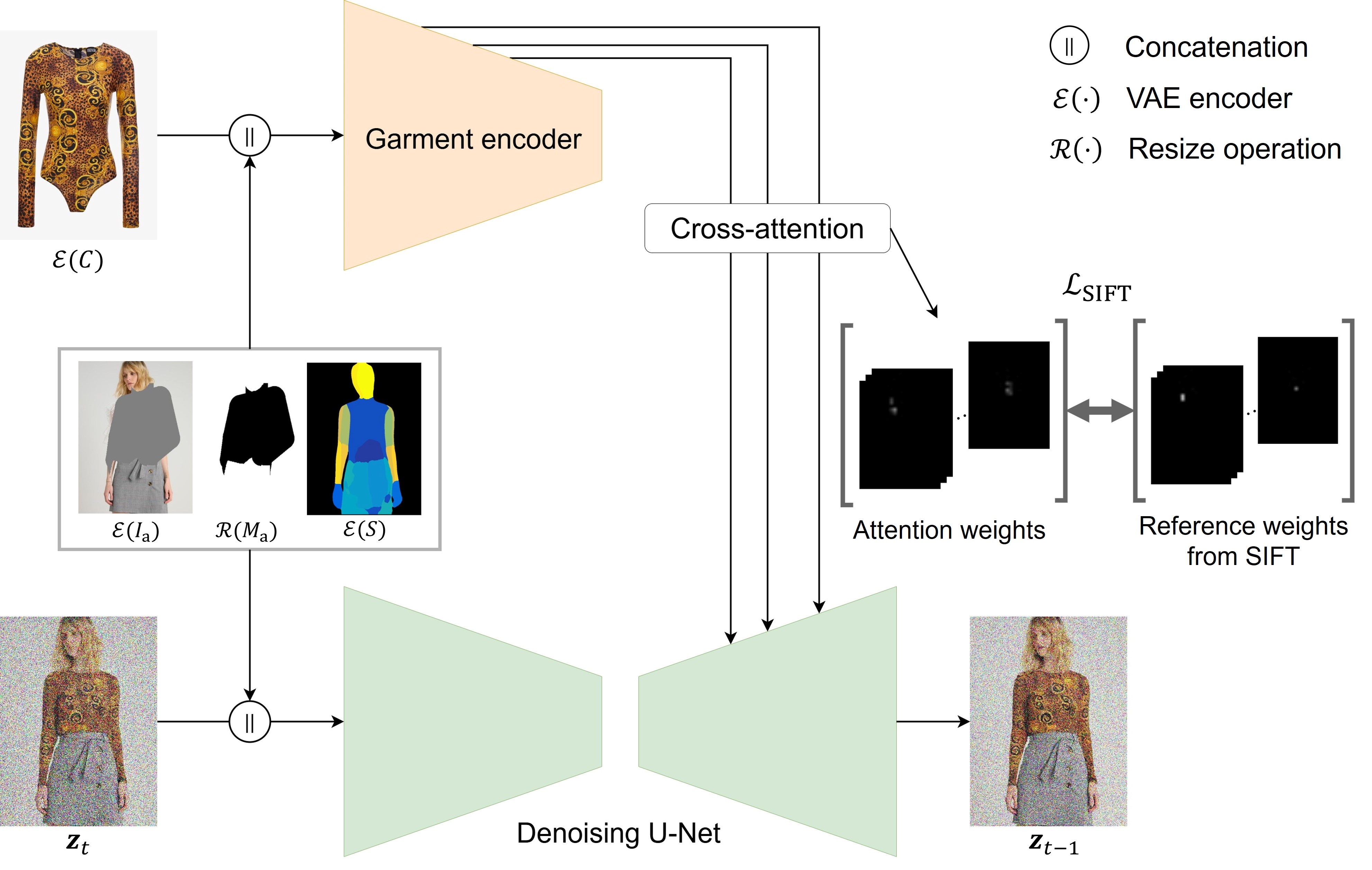}
    \caption{A diffusion step of our model. During training,
    attention weights of cross-attention are compared with SIFT-based reference attention weights.}
    \label{fig:overview}
\end{figure}
In this work, we propose utilizing Scale Invariant Feature Transform (SIFT)~\cite{sift} keypoint correspondences
to supervise the cross-attention layers of denoising U-Net in diffusion models,
so that the model learns to attend to relevant garment regions.

We build upon StableVITON~\cite{stableviton}, a state-of-the-art diffusion-based virtual try-on method,
and enhance its training by incorporating our SIFT-based cross-attention supervision loss.
To efficiently validate the effectiveness of our approach,
we fine-tune the pre-trained StableVITON model with our augmented loss function,
allowing us to isolate the contribution of the proposed SIFT supervision.
The overall architecture is illustrated in Fig.~\ref{fig:overview}.

During training, we follow the standard paired setting using paired samples:
a flattened garment image $C \in \mathbb{R}^{3 \times H \times W}$ and
a corresponding person image $I \in \mathbb{R}^{3 \times H \times W}$ showing the person wearing that garment.
We use a diffusion-based inpainting approach that
reconstructs the masked region of a garment-agnostic person representation $I_\text{a}$,
created by removing the original garment and surrounding area from $I$.
A binary mask $M_\text{a} \in \{0, 1\}^{1 \times H \times W}$ indicates which regions should remain unchanged.
For actual virtual try-on scenarios,
we use an unpaired setting where we supply a different garment $C'$ that was not originally worn in $I$.

Following recent VTON methods~\cite{ladi-vton,stableviton,ita-mdt},
our approach operates in the latent diffusion framework~\cite{ldm},
which performs the diffusion process in the compressed latent space
of a pre-trained Variational Autoencoder (VAE)~\cite{vae}.
This reduces computational costs while maintaining high-quality synthesis.
Input images are first encoded into latent representations using the encoder $\mathcal{E}(\cdot)$,
and input masks are resized to match the spatial dimensions of the latent space.
The inputs to the denoising U-Net in Fig.~\ref{fig:overview}
at diffusion time step $t$ are
the noisy latent $\boldsymbol{z}_t$, $\mathcal{E}(I_\text{a})$, $\mathcal{R}(M_\text{a})$,
and the semantic segmentation of the person's body $\mathcal{E}(S)$,
where $\mathcal{R}(\cdot)$ denotes resizing to the latent spatial resolution.
The semantic segmentation $S$ is obtained using DensePose~\cite{densepose}.

StableVITON employs a garment encoder with the same architecture as the encoder part of the denoising U-Net.
The garment encoder extracts multi-scale features from $C$,
which are fed to the corresponding cross-attention layers in the denoising U-Net based on spatial resolution.
Features at each spatial location in the garment feature maps are projected into keys and values for cross-attention,
while feature maps from the denoising U-Net are projected to queries.
Our method supervises these cross-attention layers using SIFT correspondences,
encouraging the attention weights to focus on geometrically consistent regions.

\subsection{Filtering SIFT Matches for Virtual Try-On}
To obtain reliable correspondences for attention supervision,
we apply a domain-specific filtering process to SIFT matches
between the in-shop garment and upper body region of ground truth person image.

After applying Lowe's ratio test~\cite{sift},
which retains only distinctive matches whose distances are significantly smaller
than those to second-nearest neighbors,
we further filter out mismatched keypoints by applying the following procedure tailored for virtual try-on scenarios.
First, we exploit the typical geometric constraints in virtual try-on: the person stands upright while the garment is displayed vertically.
Under these conditions, we expect small angle and scale changes between corresponding keypoints.
We therefore discard matches whose angle change or scale ratio exceeds predefined thresholds.
Second, we remove duplicate detections at identical locations,
as SIFT may output multiple keypoints with different scales or orientations at the same spatial position.
Finally, we eliminate outliers using Random Sample Consensus (RANSAC)~\cite{ransac} with a homography model.
Our filtering code will be made publicly available for reproducibility.
\begin{figure}[t]
    \centering
    \begin{subfigure}[b]{0.65\textwidth}
        \captionsetup{width=\textwidth}
        \includegraphics[width=\textwidth, keepaspectratio]{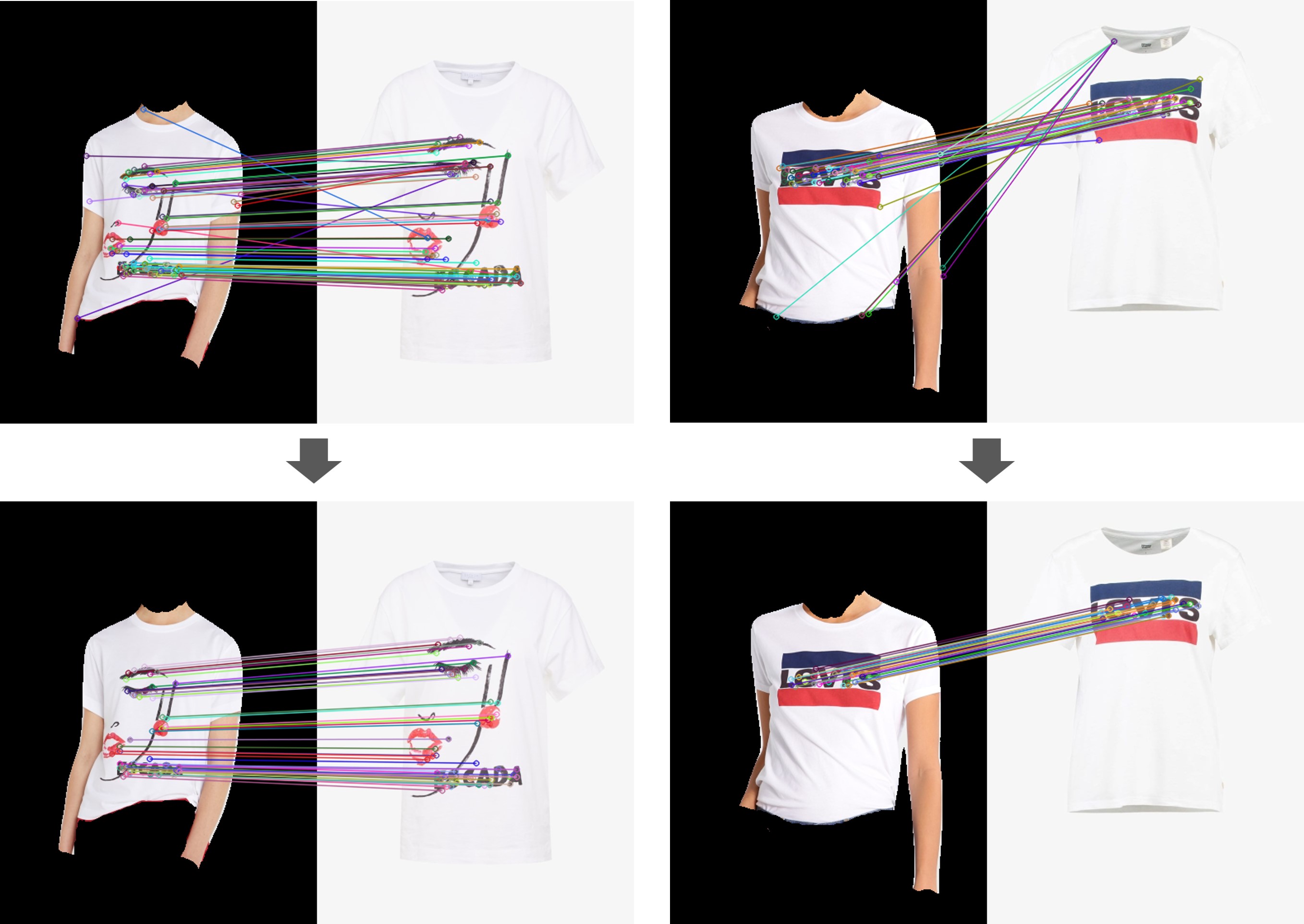}
        \caption{Two examples of filtered SIFT correspondences between garment and person images.}
        \label{fig:filtering}
    \end{subfigure}
    \hfill
    \begin{subfigure}[b]{0.3\textwidth}
        \captionsetup{width=\textwidth}
        \includegraphics[width=\textwidth, keepaspectratio]{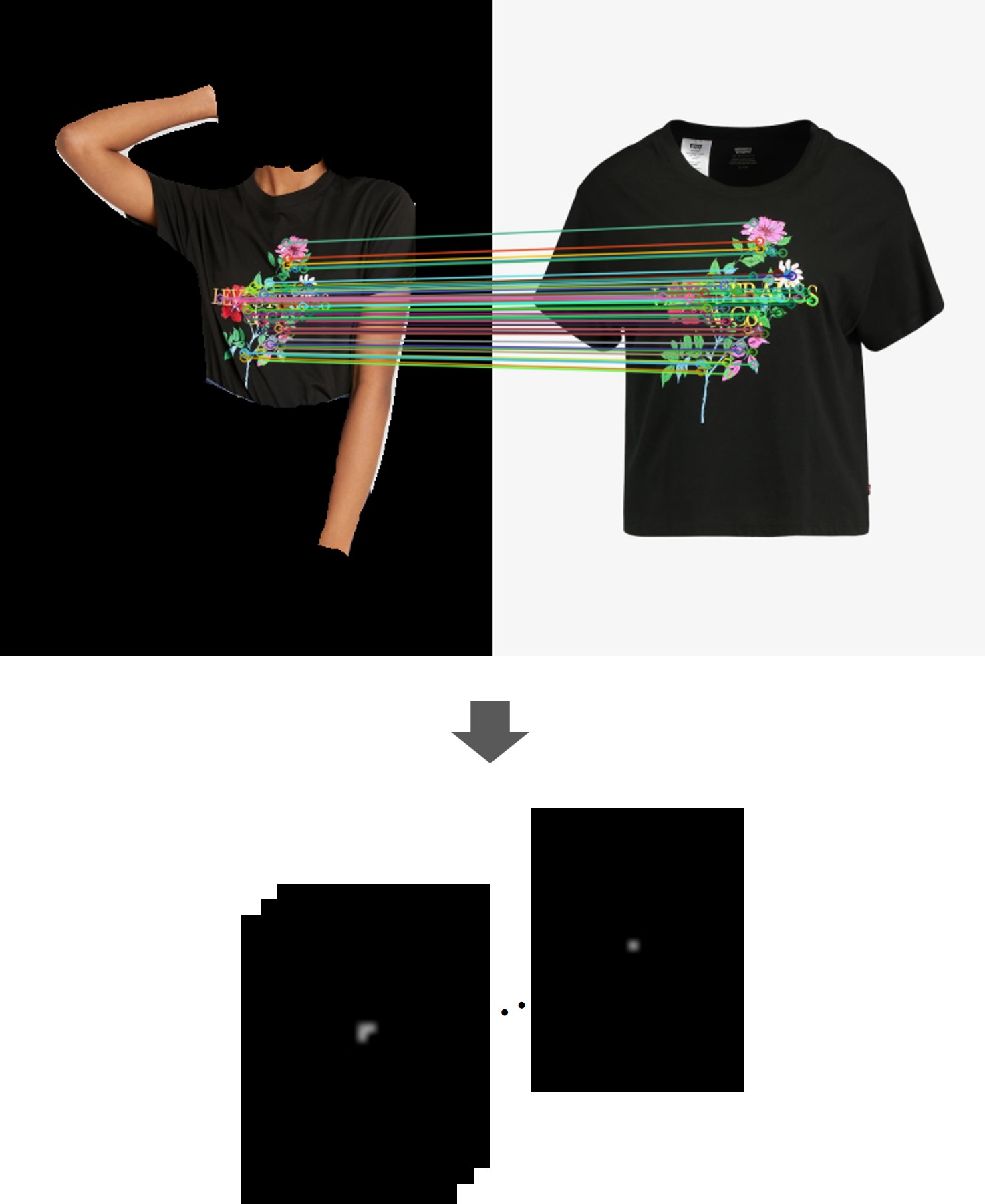}
        \caption{Converting SIFT correspondences into reference attention weights.}
        \label{fig:converting_sift}
    \end{subfigure}
    \caption{Processing SIFT correspondences for cross-attention supervision.}
    \label{fig:processing_sift}
\end{figure}

Fig.~\ref{fig:filtering} shows two examples of the filtered SIFT correspondences,
demonstrating how our filtering preserves semantically meaningful matches while removing geometric inconsistencies.
Our conservative filtering strategy prioritizes correspondence quality over quantity.
Consequently, the filtered SIFT matches are sparse like in Fig.~\ref{fig:filtering},
and no matches are found at all for plain garments without distinctive features.
This sparsity aligns with the observation
that existing VTON methods already handle simple garments well~\cite{ladi-vton,anydoor,stableviton,ita-mdt}.
This filtering process is performed as a preprocessing step before training
and does not add computational overhead during training or inference.
These filtered correspondences are then converted into supervision signals for cross-attention layers, as described next.

\subsection{Cross-attention Supervision with SIFT Correspondences}
To guide the model's attention towards relevant garment regions during reconstruction,
we introduce a novel loss term that supervises the attention weights of cross-attention layers
using filtered SIFT correspondences.

Our approach consists of three steps.
First, we convert the filtered SIFT keypoint locations from image space to
coordinates of the feature map in the garment encoder,
accounting for the resolution difference in latent diffusion models.
Second, for each query location $i$ on the person image,
we create a spatial histogram by counting corresponding SIFT keypoints
that match to location $i$, binned by their locations $j$ on the garment image.
Third, we normalize these histograms to form probability distributions
that indicate where the model should focus for each query location $i$,
resulting in reference attention weights $p_{i,j}$ (Fig.~\ref{fig:converting_sift}).

The reference attention weights serve as supervision for the cross-attention weights.
We apply cross-entropy loss to align the predicted attention weights with the target distribution
for each query location $i$ with SIFT correspondences:
\begin{eqnarray}
\mathcal{L}_{\text{SIFT}}^{l,h,i} = - \sum_{j} p_{i,j} \log(q_{l,h,i,j}), \quad i \in M_{\text{SIFT}}
\end{eqnarray}
where $M_{\text{SIFT}}$ denotes the set of query locations with at least one SIFT match,
$p_{i,j}$ is the normalized histogram value at garment location $j$ for query $i$,
and $q_{l,h,i,j}$ is the attention weight from query $i$ to key location $j$ at layer $l$ and head $h$.

The overall SIFT attention loss averages across all applicable query locations,
attention heads, and cross-attention layers:
\begin{eqnarray}
\mathcal{L}_{\text{SIFT}} = \frac{1}{L \cdot H \cdot |M_{\text{SIFT}}|} \sum_{l,h,i \in M_{\text{SIFT}}} \mathcal{L}_{\text{SIFT}}^{l,h,i}
\end{eqnarray}
where $L$ and $H$ represent the number of cross-attention layers and heads, respectively.

We integrate this supervision with the standard diffusion training objective:
\begin{eqnarray}
\mathcal{L}_t = \omega(t)||\epsilon_t - \hat{\epsilon}_t||^2_2 + \lambda_{\text{SIFT}} \mathcal{L}_{\text{SIFT}}
\end{eqnarray}
where the first term is the squared error loss for noise prediction,
$\omega(t)$ is a timestep-dependent weighting function,
and $\lambda_{\text{SIFT}}$ controls the contribution of attention supervision.
Since intermediate representations at early timesteps in the reverse process do not have clear features,
we apply SIFT supervision only when $t \leq \eta$, following~\cite{anydressing}.
Fig.~\ref{fig:overview} illustrates a diffusion step of our model,
highlighting how SIFT-based supervision is integrated into the cross-attention layers.

Note that SIFT matches are available only for the paired setting,
which is a different setting from the actual scenario of virtual try-on.
Therefore, we introduce the loss function for training supervision
rather than using the matches as additional inputs during inference.

\section{Experiments}
\subsection{Experimental Setup}
\label{ssec:experiments-setup}
We conduct experiments on VITON-HD~\cite{vitonhd},
a widely-used high-resolution virtual try-on dataset containing
\num[group-separator={,}]{11647} training pairs and \num[group-separator={,}]{2032} test pairs.
All baselines and our method are trained and evaluated on this dataset.
Following standard practice in prior work,
we resize all images to $512 \times 384$ for our experiments
and randomly select \num[group-separator={,}]{1000} pairs from the training set for validation.
We monitor validation metrics every 20 epochs and select the checkpoint
where unpaired performance peaks.
%

\subsubsection{Implementation Details}
For SIFT correspondence filtering, we set the angle threshold to $45^\circ$
and scale ratio bounds to $[0.44, 2.25]$
based on the geometric constraints of virtual try-on and empirical observations.
Our method builds upon StableVITON~\cite{stableviton},
a state-of-the-art diffusion-based virtual try-on model.
We train the network for 160 epochs using the AdamW optimizer
with a learning rate of $10^{-4}$ and batch size of 32.
For Classifier-Free Guidance (CFG)~\cite{cfg},
we omit conditioning information with a probability of 0.1 during training.
min-SNR weighting~\cite{min-snr} is used to determine the timestep-dependent loss weight $\omega(t)$.
We set the loss weight $\lambda_{\text{SIFT}}$ to 0.0005
and the diffusion timestep threshold $\eta$ to 500 for SIFT supervision.

At inference, we employ the PLMS sampler~\cite{plms} with 50 denoising steps for our model,
consistent with the StableVITON baseline.
For other diffusion models, we keep the default samplers of their official implementations,
but set 50 denoising steps for fair comparison.
Moreover, StableVITON and our method share the random seed at inference,
ensuring identical sampling noise for fair comparison.
CFG scale is set to 1.5 for our generation.
All other hyperparameters follow the default values from each baseline method's official implementation.

\subsubsection{Evaluation Metrics}
For quantitative evaluation, we employ standard metrics for both paired and unpaired settings.
In the paired setting, where ground truth images are available,
we measure reconstruction quality using
Structural Similarity Index (SSIM)~\cite{ssim} and
Learned Perceptual Image Patch Similarity (LPIPS)~\cite{lpips}.
For the unpaired setting, we evaluate the distribution similarity between real and generated images using
Fr\'echet Inception Distance (FID)~\cite{fid} and Kernel Inception Distance (KID)~\cite{kid}.
We compute SSIM and LPIPS using TorchMetrics~\cite{torchmetrics},
while FID and KID are computed using clean-fid~\cite{cleanfid}.

\subsection{Qualitative Results}
\begin{figure}[!h]
    \centering
    \includegraphics[width=0.98\linewidth]{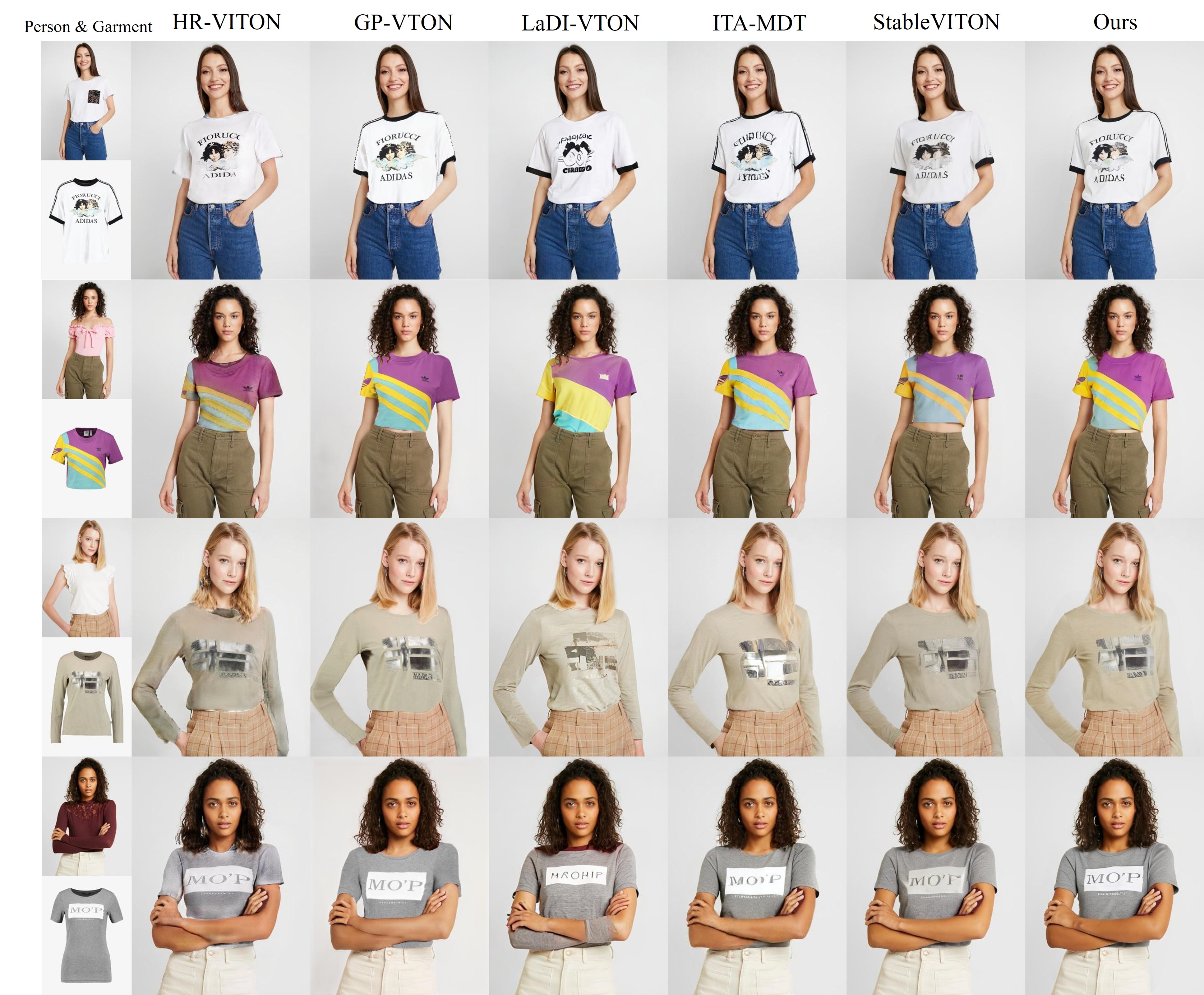}
    \caption{Qualitative comparisons with the unpaired setting. Best viewed when zoomed in.}
    \label{fig:unpair_comp}
\end{figure}
Fig.~\ref{fig:unpair_comp} presents qualitative comparisons between our method and baselines.
HR-VITON~\cite{hrviton} and GP-VTON~\cite{gpvton} are GAN-based explicit warping methods,
and other methods are based on diffusion models with implicit warping,
which reconstructs given garments via attention mechanisms.
In these examples, GP-VTON retains garment details well by directly copying warped garment regions,
but suffers from warping artifacts, such as lack of natural wrinkles.
LaDI-VTON~\cite{ladi-vton} struggles to maintain given garment identities.
ITA-MDT~\cite{ita-mdt} reconstructs garment features better,
but fine details, such as logos and illustrations, are unstable and have low fidelity.
StableVITON~\cite{stableviton} and our method preserve more garment details,
but StableVITON often omits peripheral garment features,
such as the sleeve trim in the first row and the bottom yellow stripe in the second row.
In contrast, our SIFT-based supervision better reconstructs such features,
which StableVITON tends to overlook.
Moreover, our method consistently improves clarity and sharpness throughout these examples,
as visible in the text rendering (rows 1, 4),
stripe completeness (row 2), and graphic details (row 3).

\subsection{Quantitative Results}
\begin{table}[!h]
    \centering
    \captionsetup{width=.75\textwidth}
    \caption[hoge]{Quantitative comparison on VITON-HD test set. KIDs are multiplied by 1000 for better readability.}
  \begin{adjustbox}{width=0.6\linewidth,center}
    \begin{tabular}{ l c c c c } \hline
Method &  SSIM↑ & LPIPS↓ & $\text{FID}_{\text{u}}$↓ & $\text{KID}_{\text{u}}$↓ \\ \hline
HR-VITON~\cite{hrviton} & 0.879 & 0.0970 & 10.845 &  2.434 \\
GP-VTON~\cite{gpvton} & \textbf{0.892} & 0.0831 &  9.611 & 1.436 \\
LaDI-VTON~\cite{ladi-vton} & 0.873 & 0.0940 & 9.414 & 1.660 \\
StableVITON~\cite{stableviton} & \underline{0.891} & \underline{0.0752} & \underline{9.304} & \underline{1.397} \\ \hline
SIFT-VTON (Ours) & 0.889 & \textbf{0.0738} & \textbf{8.593} & \textbf{0.754}  \\ \hline
    \end{tabular}
  \end{adjustbox}
  \label{tab:quantitative}
\end{table}
Table~\ref{tab:quantitative} presents quantitative comparisons
\footnote{ITA-MDT is excluded from quantitative comparisons as it trains at 512×512 resolution.}
with state-of-the-art methods on the VITON-HD test set.
SIFT-VTON achieves the best unpaired metrics,
with a 7.6\% $\text{FID}_{\text{u}}$ reduction (9.304 → 8.593)
and a 46.0\% $\text{KID}_{\text{u}}$ reduction (1.397 → 0.754) compared to StableVITON.

For paired metrics, our method achieves comparable performance,
with slightly improved LPIPS (0.0738 vs 0.0752) and slightly lower SSIM (0.889 vs 0.891).
The substantial improvements in FID and KID demonstrate that
SIFT-based supervision enhances generation quality in the unpaired setting.
The unpaired setting better reflects the actual virtual try-on use case,
where VTON systems cannot rely on residual garment patterns from the input person image.

These improvements validate that explicit geometric correspondence supervision
helps the model learn more accurate spatial alignment.
The results are consistent with qualitative observations of improved text clarity,
pattern completeness, and peripheral feature preservation.

\subsection{Attention Visualization}
\begin{figure}[!h]
    \centering
    \begin{subfigure}[t]{0.98\textwidth}
        \captionsetup{width=\textwidth}
        \includegraphics[width=\textwidth, keepaspectratio]{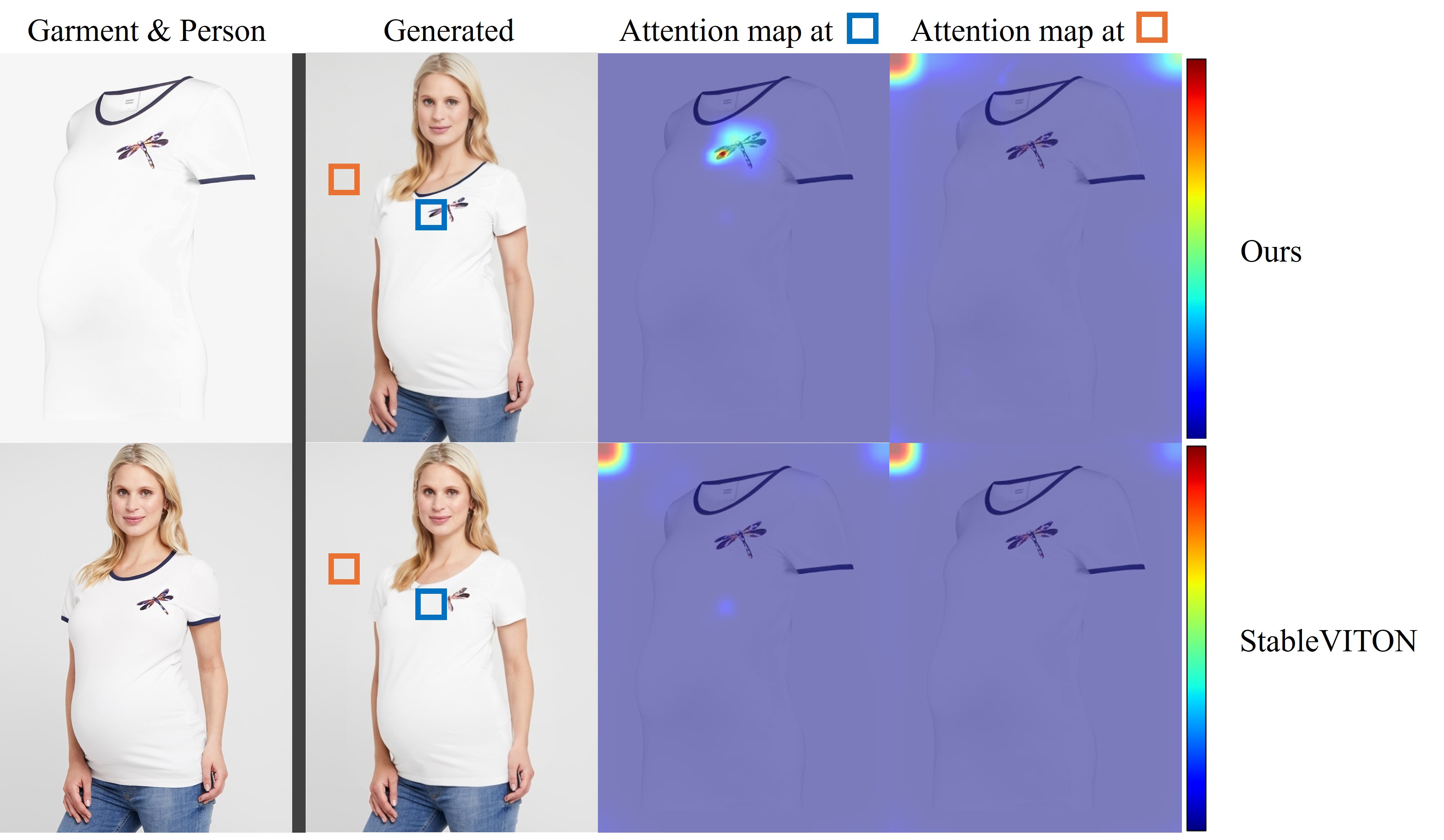}
        \caption{Dragonfly graphic example.}
        \label{fig:attn_vis_sub}
        \vspace*{0.3cm}
    \end{subfigure}
    \begin{subfigure}[t]{0.98\textwidth}
        \captionsetup{width=\textwidth}
        \includegraphics[width=\textwidth, keepaspectratio]{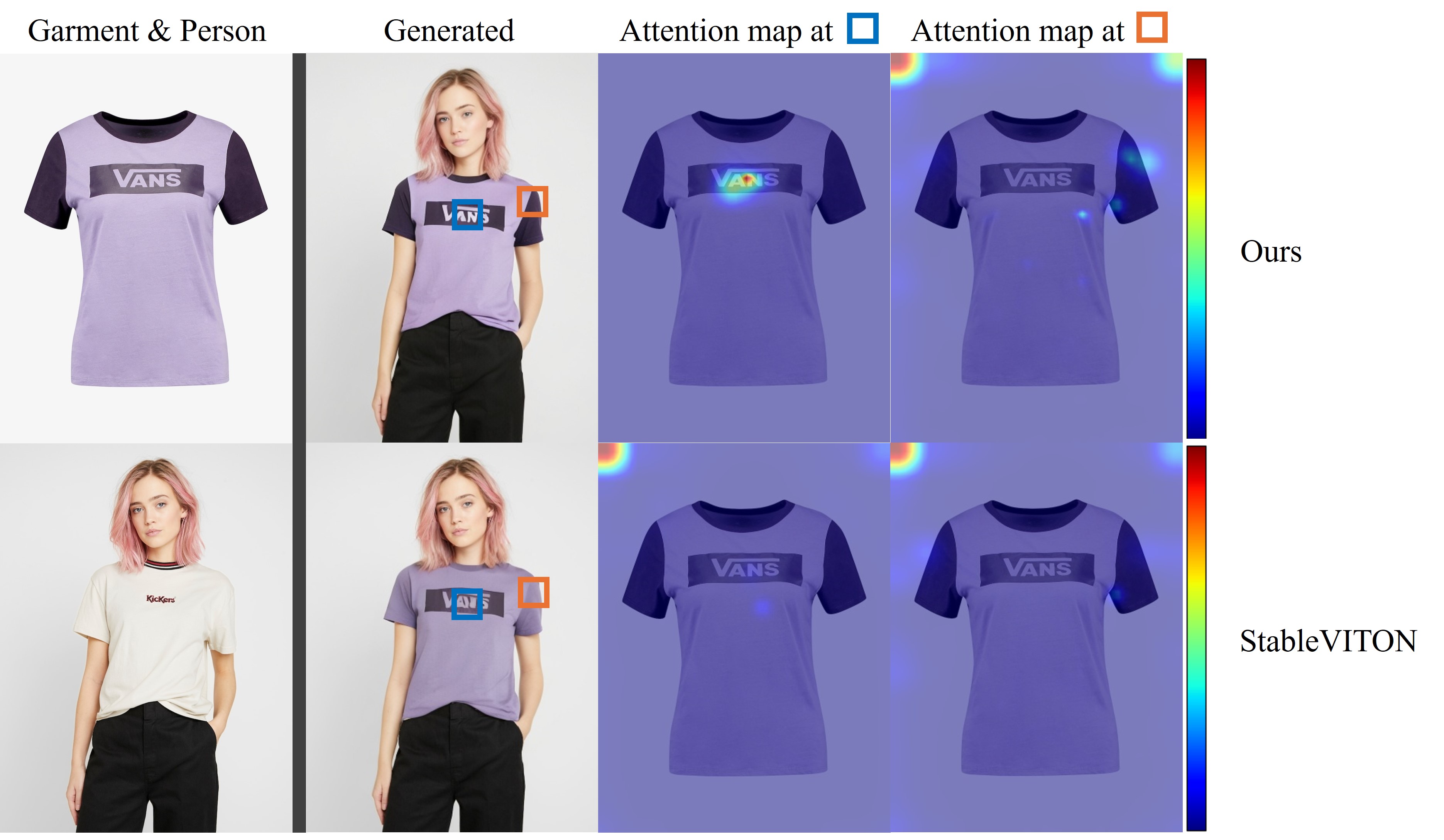}
        \caption{Text logo example.}
        \label{fig:attn_vis_sub_2}
    \end{subfigure}
    \caption{Visualization of cross-attention maps from StableVITON and our method.}
    \label{fig:attn_vis}
\end{figure}
To understand how SIFT supervision affects learned attention patterns,
we visualize cross-attention maps for query locations on generated images.
Fig.~\ref{fig:attn_vis} shows attention maps
for query locations marked by colored boxes.
The maps are averaged across cross-attention layers and attention heads at timesteps $t < \eta = 500$,
where SIFT supervision is applied during training.

Our method produces significantly more focused attention on corresponding garment features.
In Fig.~\ref{fig:attn_vis_sub}, for the dragonfly graphic query (blue box),
attention sharply concentrates on the dragonfly.
Similarly, in Fig.~\ref{fig:attn_vis_sub_2}, attention precisely targets the text logo (blue box).
For sleeve regions (orange box in Fig.~\ref{fig:attn_vis_sub_2}),
both methods show similar attention patterns,
though our method concentrates more on the corresponding sleeve area.
In contrast, StableVITON exhibits highly diffuse attention in both examples.
For non-garment query locations such as background regions,
both methods attend to the upper corners of the garment feature map
(orange box in Fig.~\ref{fig:attn_vis_sub}).

These consistent patterns across different feature types,
including graphics and text, demonstrate that
SIFT-based supervision successfully guides the model to learn spatially precise attention.
This directly leads to improved detail clarity and preservation in generated images.
\section{Conclusion}
We introduced SIFT-based cross-attention supervision for diffusion-based virtual try-on,
providing explicit geometric guidance to cross-attention.
Our method converts filtered SIFT correspondences into probability distributions
that supervise cross-attention layers during training,
enabling more precise spatial alignment of garment features.

Experiments on the VITON-HD dataset demonstrate significant improvements in generation quality
while maintaining competitive reconstruction performance.
Qualitative results show superior preservation of fine details
including text clarity and pattern alignment, compared to state-of-the-art methods.
Attention visualizations confirm that our supervision produces
more focused attention on relevant garment regions.

This work demonstrates that classical geometric methods like SIFT
can effectively guide modern diffusion models,
and that explicit geometric supervision is complementary to
architectural innovations in diffusion-based virtual try-on.
The approach opens possibilities for combining
explicit correspondence estimation with diffusion models
in other conditional synthesis tasks.

\subsubsection{Limitations}
Although our SIFT correspondence filtering is conservative,
some mismatches still remain,
potentially introducing noise into the supervision signal.
Additionally, the loss weighting hyperparameters ($\lambda_{\text{SIFT}}$ and $\eta$)
are set based on validation performance without exhaustive grid search;
more systematic hyperparameter optimization could yield further improvements.


%
%

\subsubsection{Acknowledgements}
This work was partially supported by JSPS KAKEN Grant Numbers 21K11967 and 24K15012.

%
%
%
\bibliographystyle{splncs04}
\bibliography{refs}

\end{document}